\documentclass[runningheads]{llncs}
\usepackage{float} % Required for the [H] option

\usepackage[T1]{fontenc}
\usepackage{graphicx}
\usepackage{amssymb} 
\usepackage{amsmath}
\usepackage{xcolor}

\begin{document}
\title{Symbolic Graph Inference\\for Compound Scene Understanding}
\author{FNU Aryan\inst{1} \and
Simon Stepputtis\inst{1} \and
Sarthak Bhagat\inst{1} \and
Joseph Campbell\inst{1} \and
Kwonjoon Lee\inst{2} \and
Hossein Nourkhiz Mahjoub \inst{2} \and
Katia Sycara \inst{1}}

\authorrunning{Aryan et al.}
\institute{Carnegie Mellon University \and
Honda Research Institute
}
\maketitle              % typeset the header of the contribution
\begin{abstract}
Scene understanding is a fundamental capability needed in many domains ranging from question-answering to robotics. 
Unlike recent end-to-end approaches that must explicitly learn varying compositions of the same scene, our method reasons over their constituent objects and analyzes their arrangement to infer a scene's meaning.
We propose a novel approach that reasons over a scene's scene- and knowledge-graph, capturing spatial information while being able to utilize general domain knowledge in a joint graph search.
% Our approach formulates scene understanding as identifying the combination of its basic components.
Empirically, we demonstrate the feasibility of our method on the ADE20K dataset and compare it to current scene understanding approaches.

\end{abstract}
\section{Introduction}
\vspace{-10pt}
Scenes understanding is crucial for numerous applications, including, but not limited to path planning for robotic agents \cite{lundeen2019autonomous} and developing assistive human companions \cite{li2023airloc}.
However, current scene understanding approaches \cite{cimpoi2015deep} often interpret them as indivisible entities, overlooking their intricate composition. However, such scenes \cite{zhou2017scene} are not merely a singular entity, but rather the sum of their parts.
For instance, consider a \textit{kitchen}: while its composition can vary, e.g., assume a \textit{kitchen} without an \textit{oven}, it is still a \textit{kitchen}, given that other descriptive components (e.g., \textit{stove, sink, fridge, ...}) are still present.
On the other hand, consider a harbor without its defining element - \textit{water} -, as shown in Fig.~\ref{qualitative}, which should not be identified as a harbor despite the presence of boats.
% This illustrates how the essence of an entity lies in its parts, rather than in its indivisible whole.

\begin{figure}[t]
    \centering
    \includegraphics[width=\textwidth]{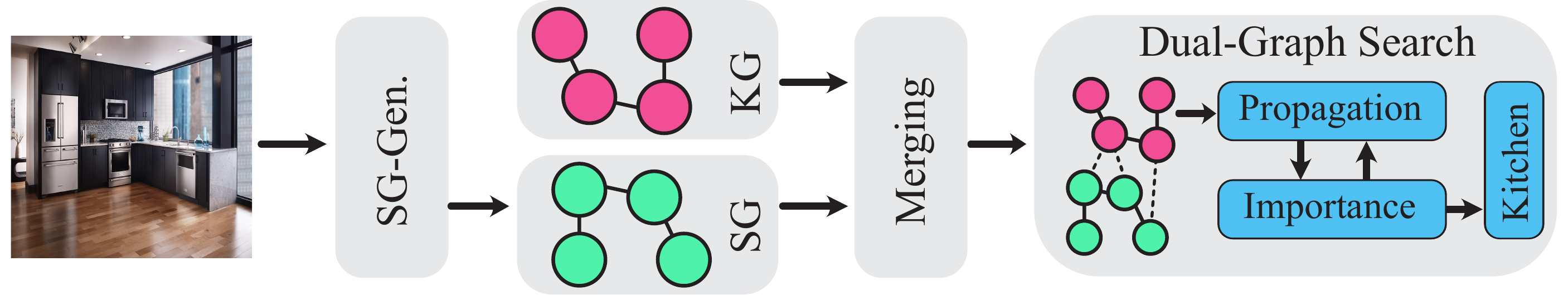}
    \caption{The overview of our multi-graph reasoning approach which uses both the scene and knowledge graph and jointly reasons over them.} 
    \label{approach}
    \vspace{-15pt}
\end{figure}

% These complex entities can be considered compound concepts comprised of various individual concepts, linked together through a knowledge graph.
One way of capturing the constituent elements of a compound scene is through a Knowledge Graph (KG), linking base elements to their compound interpretation.
For this work, we consider such complex scenes as compound concepts (e.g., \textit{kitchen}) made up of multiple primitive constituent concepts (e.g., \textit{stove} or \textit{fridge}). 
% Neuro-symbolic architectures, leveraging symbolic knowledge graphs, offer valuable tools for unraveling the intricacies of these compound concepts.
However, a KG does not capture all the necessary information to reason over such scenes.
For example, in Fig.~\ref{qualitative}, the existence of a \textit{dog} and \textit{sled} could be seen as the compound concept of \textit{dogsled}, but the spatial relationships between the basic entities and their quantities play an important role.  
% Understanding the specific relationships between objects, such as a dog being in front of the sled to pull it or a dog sitting on the sled, is crucial.
To this end, we propose to combine scene and knowledge graphs, harnessing the ability of Scene Graphs (SG) to account for multiple instances of the same concept as well as their spatial relationships, while knowledge graphs allow for high-level reasoning for scene understanding. 
To this end, we propose a novel approach (see Fig.~\ref{approach}) that dynamically establishes a link between the spatial information in the scene graph and the domain knowledge encoded in the knowledge graph, before conducting a joint search over the combined domain.  
% Our multi-graph graph search approach merges both the scene graph and the knowledge graph, enabling joint reasoning in the combined domain. 
% This integration is beneficial for various downstream tasks like compound concept prediction.
% To elucidate our contributions, we introduce a graph merging module that consolidates scene and knowledge graphs, enabling seamless joint reasoning.
Additionally, we leverage techniques such as dynamic graph propagation to improve the runtime and scalability of the method. Our main contributions include:
\begin{itemize}
    \vspace{-10pt}
    \item We propose a novel dual-graph graph search approach that jointly reasons over scene and knowledge graphs.
    \item In comparison to prior work, we utilize a dynamic exploration approach that automatically determines when enough information has been considered.
    % \item Through experimental validation, demonstrating the superiority of our method over baselines like GPT4-Vision \cite{openai2024gpt4} and previous neuro-symbolic methods, including knowledge graph-only approaches.
    \item In initial experiments, we demonstrate the feasibility of our approach compared to symbolic-only and neural-only approaches. 
\end{itemize}

% \begin{figure}[t]
% \includegraphics[width=\textwidth]{Figs/Multi_graph GSNN Without Sction Anticipation.pdf}
% \caption{The overview of our multi-graph reasoning approach which uses both the scene and knowledge graph and  jointly reasons over them} \label{approach}
% \end{figure}

\vspace{-15pt}
\section{Methodology}
\vspace{-10pt}

In this work, we aim to identify compound concepts in images. Formally, we take an input image $\mathbf{I}$ and domain KG $\mathcal{K}$ to classify the set of concepts and compound concepts $\mathcal{C}$ through our proposed method $\mathcal{C} = f_\Theta(\mathbf{I}, \mathcal{K})$, where $\theta$ reflects the trainable parameters of our neural components. Our approach works in three steps: 1) creating a scene and knowledge graph from input data, 2) merging the graphs, and 3) searching over the merged graph and doing our final prediction (see Figure~\ref{approach}).

\textbf{Scene Graph Generation.}
To generate a scene graph $\mathcal{S}$, we utilize an object detector, namely Faster R-CNN~\cite{Girshick_2015_ICCV}, to detect initial concepts $\hat{\mathbf{c}}$ in the provided scene. 
These objects serve as concepts in our SG, while edge-types (encoding the spatial relationships between the nodes) are predicted by a small neural network $\mathbf{A}_S = f^{\text{EDGE}}_\theta(\hat{\mathbf{c}})$ inspired by ~\cite{zhuang2017towards}.
% In the scene graph $\mathcal{S}$, objects serve as nodes, and their spatial relations serve as directional edges represented by adjacency matrix $\mathbf{A_S}$.
% We employ scene-specific object embeddings to represent each node within the graph. These 768-dimensional embeddings are generated by applying a ViT-based object encoder on object bounding boxes obtained using Faster R-CNN \cite{Girshick_2015_ICCV}.
Each node $\hat{c}_i \in \hat{\mathbf{c}}$ in the SG is represented as an embedding derived from its bounding box through ViT~\cite{DBLP:journals/corr/abs-2010-11929}.
% Additionally, we determine spatial relations between objects using a relation prediction network inspired by \cite{zhuang2017context} which infers the relationships between pairs of objects based on their positions. 
% We utilize a hand-crafted knowledge graph \(G\) as a structured source of domain knowledge. We use bi-direction edges in the graph to connect various concepts to the relevant compound concepts represented by adjacency matrix $\mathbf{A_G}$. The nodes of the Knowledge Graph are represented using 100-dimensional GloVe embeddings.

% Later, we detail the Graph Merging Module, which unifies disparate graphs. Subsequently, we introduce the Merged Graph Search Network for reasoning about the merged graph.

\textbf{Graph Merging Module.}
The scene graph $\mathcal{S}$ encapsulates scene-specific information, while the knowledge graph $\mathcal{K}$ holds general domain knowledge of how individual concepts are related to compound concepts and their affordances. Our KG is hand-designed for our particular use case, inspired by the concepts present in the ADE20K dataset.
The goal of our graph-merging approach is to combine the spatial information of the SG with the domain knowledge of the KG.
% We explore different initialization schemes for scene graph nodes crucial in graph merging. The exhaustive approach activates all nodes, even redundant ones, which may introduce noise. Alternatively, random initialization selects nodes and their neighbors randomly. We enhance randomness with a two-stage approach: first, selecting a node and activating its neighbors, then repeating the process for another node not previously selected.

% Subsequently, by connecting a subset (potentially random or all) of nodes between the graphs as explained later, we formed a merged graph ($M$) representing a unified knowledge space. 

% In the graph merging process, we first generate the SG as described previously and initialize the KG from the set of detected concepts $\hat{\mathbf{c}}$, forming our merged graph $\mathcal{M}$.
% % Connections are formed by dynamically linking the same SG and KG nodes.
% % % An intuitive way is to exhaustively initialize nodes, potentially including redundant ones. 
% % We start graph-search from a randomly chosen node in the SG part of our merged graph $M$.
% We randomly select a node and its neighbors in the Scene Graph (SG) to initiate the graph search (see next paragraph for details on the search). 
% We refer to this ``partially active graph'' over $\mathcal{M}$ as $\mathcal{M}^A$.
% Connections are formed and dynamically updated by linking shared nodes between the active SG and Knowledge Graph (KG). 
In this process, we first generate the SG as described previously and initialize the KG from the set of detected concepts $\hat{\mathbf{c}}$.
We then randomly select a node and its neighbors in the SG and form connections by linking shared nodes between the SG and KG to generate our merged graph $\mathcal{M}$ containing entire SG and KG. The sub-graph consisting of connected SG and KG nodes is referred to as ``active graph'' $\mathcal{M}^A$ over $\mathcal{M}$, and it is used to initiate the graph search (see next paragraph for details on the search). 
% \textcolor{red}{[Check the previous sentence. It is unclear what the difference between $M$ and $M^A$ is.]}
% Graph search starts from these initialized nodes in our merged graph $\mathcal{M}$. 
After our entire graph search is completed (see method below) and SG nodes that have not been activated remain, we conduct an additional search starting from such previously non-activated nodes. 
Through this approach, we ensure the entire scene has been considered, even if comprised of multiple compound scenes (e.g., having a \textit{kitchen} on the right, and a \textit{living room} on the left of an image). 

\textbf{Merged Graph Search Network.}
% The Graph Search Neural Network (GSNN) operates over a merged graph ($M$), comprising two key components: the \textit{Propagation Network} and the \textit{Importance Network}. We draw inspiration from our prior research for these components.
% [Simon's Proposal]:
Having formed the initial connection between SG and KG, we conduct a joint graph search over the active merged graph $\mathcal{M}^A$.
Inspired by~\cite{bhagat2023sample}, we leverage a three-staged approach: 1) we utilize a \textit{propagation network}, which calculates neighborhoods of the currently considered concepts; 2) an \textit{importance network}, which decides which concepts should be expanded; and 3) a \textit{task-based classifier} that predicts final output.
In contrast to~\cite{bhagat2023sample}, we utilize a dynamically determined number of iterations between the propagation- and importance-networks:
\vspace{-2pt}
\textit{Propagation Network:}
% The Propagation Network updates active node states within the active graph \( C \) and outputs \( O' \in {R}^{P \times F} \), where \( P \) represents the number of active nodes and \( F \) denotes the node feature size. It leverages adjacency matrix \( A_M \) to encode neighborhood information. The update equation is give by: \( H^{(t+1)} = \sigma(A_MH^{(t)}W) \) Where \( H^{(t)} \) is the hidden states of nodes at iteration \( t \), \( \sigma \) is the activation function, and \( W \) is the weight matrix.
The propagation network updates active graph $\mathcal{M}^A_{t}$ in each interaction $t \in T$  and outputs an updated active graph $\mathcal{M}^A_{\hat{t}}$ in which current neighbors in graph $\mathcal{M}^A_{t}$ are considered where $\hat{t}$ indicates a partial iteration $t$.

% \( \mathbf{O^t} \in {\mathbb{R}}^{P \times F} \), where \( P \) represents the number of active nodes and \( F \) denotes the node feature size and $t$ is iteration. It utilizes the adjacency matrix \( \mathbf{A_M} \), allowing nodes to exchange information and refine their representations based on connections within the graph structure.

% \textbf{Importance Network:}
% The Importance Network operates in alternation with the Propagation Network across \( T \) cycles determined dynamically, determining whether adjacent nodes to currently active ones should be activated. This step is crucial, as indiscriminate node expansion at every step could become computationally burdensome for large graphs. The importance \( v_t^n \) of each node at iteration \( t \) is calculated conditioned on the global image context \( e_I \) as follows:
% \[ v_t^n = f_I(h_t^n, e_I) \]
% \( f_I(\cdot) \) denotes a multi-layer perceptron (MLP). Nodes with importance scores above a certain threshold \( \gamma \) are activated for the subsequent propagation cycle.
\vspace{-5pt}
\textit{Importance Network:}
The importance network works alongside the propagation network in an iterative manner, determining which adjacent nodes to $\mathcal{M}^A_{\hat{t}}$ should be added to $\mathcal{M}^A_{t+1}$ in the next step. At each iteration, the importance of each adjacent node is computed based on the set of current nodes as well as the overall input image. If a node exceeds a pre-defined importance threshold $\gamma$, it is added to $\mathcal{M}^A_{t+1}$. 
Through this process, intuitively, the utility of additional concepts, such as compound concepts, are explicitly evaluated and searched for, given the currently active concepts and the whole image context.

% \textbf{Final Classifier:} The third and final step involves the classification of the active nodes \( P \subseteq N \) in the input image \( I \), where \( N \) represents the set of all possible nodes. These nodes are determined from the state representations 
%  \(O_{final}\) of all the expanded nodes in the active graph 
% \( P \), along with the global image embedding \( e_I \) . Utilizing a single fully connected layer, a probability distribution over all the nodes is predicted as :
% \( c = f_C(O_{final}, e_I) \). 
% In order to facilitate interpretation by a human user, we also provide the graph of active nodes \( P \), thereby offering insights into why certain classifications may have been made.
% \vspace{-5pt}
\textit{Task-based classifier:} 
% The final step involves classifying the active nodes \( P \subseteq N \), where \( N \) represents all possible nodes. These nodes are classified using the state representations \( \mathbf{O_{final}} \) of nodes in the active graph \( M_P \), along with the global image embedding \( e_I \). Using a single fully connected layer, a probability distribution over all nodes is predicted: \( c = f_C(\mathbf{O_{final}}, e_I) \). 
% Additionally, we provide the graph of active nodes \( P \) for human interpretation, offering insights into classifications.
After $T$ iterations, we employ a linear classifier over $\mathcal{M}^A_T$ to determine overall concepts that should be active for a scene. Intuitively, unlike importance networks, this conducts a final filter that can remove nodes.  

 \textbf{Dynamic Propagation:}
 % Unlike fixed-step methods, our approach optimizes runtime efficiency and ensures graph stability by dynamically setting propagation steps.
 Besides the joint graph search, one of our main contributions is the dynamic selection of $T$.
 % In our work, the propagation and importance network work in unison through an iterative process. 
 Through our dynamic approach, we halt further expansion (i.e., further iterations) if no additional nodes 
 have been added in the previous iteration or if no added node exceeds an importance of $\lambda$. 
 Through this, we allow the network to propagate and include new information as long as it is deemed important enough while simultaneously saving additional iterations by setting a threshold $\lambda$.
 We tune $\lambda$ and set it to $0.75$, which significantly reduces the runtime while retraining high performance.

\section{Experimental Results}
\vspace{-5pt}
% In this section, we comprehensively evaluate our approach.
% In this section, we demonstrate our methods' performance on compound concept prediction. We will first discuss about the dataset and evaluation metrics used in the experiments and outline baseline models for comparison. Through comparison, we highlight our model's strengths and showcase its efficacy with qualitative examples and conduct an ablation study to analyze key components' contributions.

In this section, we evaluate the performance of our method on compound concept prediction in scene understanding and compare it with a set of relevant baselines, both using symbolic, or neural architectures.  
% We discuss the dataset, evaluation metrics, and baseline models used. 
% Further, we demonstrate its efficacy qualitatively and conduct an ablation study to analyze key components' contributions.

\textbf{Dataset and Evaluation Metrics.}
% \textbf{Write bout previous datasets Places 365 why are we not using them}
The ADE20K dataset \cite{zhou2017scene} comprises high-resolution images sourced from the Places 365 and SUN datasets. It consists of ground-truth object labels (concepts) along with overall scene labels (compound concepts).
For our study, we selected scene categories containing over $100$ images. To enhance dataset clarity, we removed certain ambiguous classes and merged them with their parent class. For instance, classes like \textit{Mountain Snowy} were merged into broader category \textit{Mountain}, while \textit{Attic} was consolidated under \textit{Bedroom}. With these preprocessing steps, our experiments involved $20$ classes. 
We utilize top-1 accuracy to compare the efficacy of our approach.

\textbf{Compound Concept Prediction.}
Our model's performance is evaluated against two types of baselines: object-level and image-based baselines. \textbf{Object-level} baselines do not use the whole image representation and only use the objects extracted from the image. These baselines include the KG baseline, where all detected concepts in a particular image are passed to the KG for compound concept prediction. The GPT baseline employs GPT3.5 and GPT4 \cite{openai2024gpt4}, taking active concepts as input and predicting compound concepts among all given compound concepts. Additionally, the Human baseline reflects human accuracy in classifying compound concepts given a set of active concepts. Our approach on the object level does not use Image conditioning in Importance Network.
% \begin{table}[H]
% \begin{minipage}[t]{0.5\linewidth}
% \centering
% \footnotesize
% \begin{tabular}{|l|c|c|}
%     \hline
%     Method (Object) &  Subset & Test set  \\
%     \hline
%     Chat GPT 3.5 & 49\% & 43.71\% \\
%     ChatGPT 4 & 65\% & 61.89\% \\
%     Knowledge Graph & 57\% & 65.34\% \\
%     Human Baseline & 81\% & NA \\
%     \textbf{Ours} & 66\% & 74.52\% \\
%     \hline
% \end{tabular}
% \caption{Object based methods}
% \label{tab:perfromance_object}
% \end{minipage}%
% \begin{minipage}[t]{0.5\linewidth}
% \centering
% \footnotesize
% \begin{tabular}{|l|c|c|}
%     \hline
%     Method (Image) & Subset & Test set \\
%     \hline
%     ViT Baseline & 82\% & 85.82\% \\
%     % Ours (Without SG relations) & 94\% & 96.38\% \\
%     GPT-4 Vision & 96\% & NA \\
%     Human Baseline & 97\% & NA \\
%     \textbf{Ours}  & 94\% & 96.25\% \\
%     \hline
% \end{tabular}
% \caption{Image Based Methods}
% \label{tab:performance_image}
% \end{minipage}
% \end{table}
% \vspace{-\baselineskip}
\vspace{-12pt}
\begin{table}[h]
\begin{minipage}[t]{0.5\linewidth}
\centering
\footnotesize
\begin{tabular}{lcc}
    \hline
    Method (Object) &   Subset   & Test set  \\
    \hline
    Knowledge Graph & 57\% & 65.34\% \\
    GPT3.5 & 49\% & 43.71\% \\
    GPT4 & 65\% & 61.89\% \\
    \textbf{Ours} & 66\% & 74.52\% \\
    Human Baseline & 81\% & NA \\
    \hline
\end{tabular}
\caption{Object-based methods}
\label{tab:perfromance_object}
\end{minipage}%
\begin{minipage}[t]{0.5\linewidth}
\centering
\footnotesize
\begin{tabular}{lcc}
    \hline
    Method (Image) & Subset & Test set \\
    \hline
    ViT & 82\% & 85.82\% \\
    % Ours (Without SG relations) & 94\% & 96.38\% \\
    GPT4-Vision & 96\% & NA \\
    \textbf{Ours}  & 94\% & 96.25\% \\
    Human Baseline & 97\% & NA \\
    \hline
\end{tabular}
\caption{Image-based Methods}
\label{tab:performance_image}
\end{minipage}
\end{table}
\vspace{-\baselineskip}
\vspace{-\baselineskip}
On the other hand, \textbf{image-level} baselines involve end-to-end image input. For the VIT \cite{DBLP:journals/corr/abs-2010-11929} baseline, we finetune a pretrained transformer for compound concept classification. The human baseline represents human accuracy in classifying compound concepts given the image. We also utilize GPT4-Vision as a baseline, where it is provided with an image input and prompted to classify the compound concept among all given compound concepts. These baselines provide a comprehensive evaluation of our model's performance across different input modalities and methodologies.

In consideration of the computational cost associated with running inference on GPT4-Vision, we partition our total test set, comprising $2200$ images, into a subset of $100$ images. This allows a fair comparison with GPT4-Vision. Each compound concept is represented by $5$ images in our smaller test set.

% \begin{table}[htbp]
%     \centering
%     \begin{tabular}{|l|c|c|}
%         \hline
%         Method (Object based) & Test Subset & Complete Test dataset \\
%         \hline
%         Chat GPT 3.5 & 49\% & 43.71\% \\
%         ChatGPT 4 & 67\% & 61.89\% \\
%         Knowledge Graph only & 57\% & 65.34\% \\
%         SG + KG (Ours) & 66\% & 74.52\% \\
%         Human Baseline & 81\% & NA \\
%         \hline
%     \end{tabular}
%     \caption{Performance Comparison (Object based methods)}
%     \label{tab:perfromance_object}
% \end{table}
% \vspace{-\baselineskip}
% \vspace{-\baselineskip}
% \vspace{-\baselineskip}

% \begin{table}[htbp]
%     \centering
%     \begin{tabular}{|l|c|c|}
%         \hline
%         Method (Image Based) & Test Subset & Complete Test Dataset \\
%         \hline
%         ViT Baseline & 82\% & 85.82\% \\
%         % Ours (Without SG relations) & 94\% & 96.38\% \\
%         Ours  & 94\% & 96.25\% \\
%         GPT-4 Vision & 96\% & NA \\
%         Human Baseline & 97\% & NA \\
%         \hline
%     \end{tabular}
%     \caption{Performance Comparison (Image Based Methods)}
%     \label{tab:performance_image}
% \end{table}
% \vspace{-\baselineskip}

\begin{figure}[t]
\centering
\includegraphics[width=0.9\textwidth]{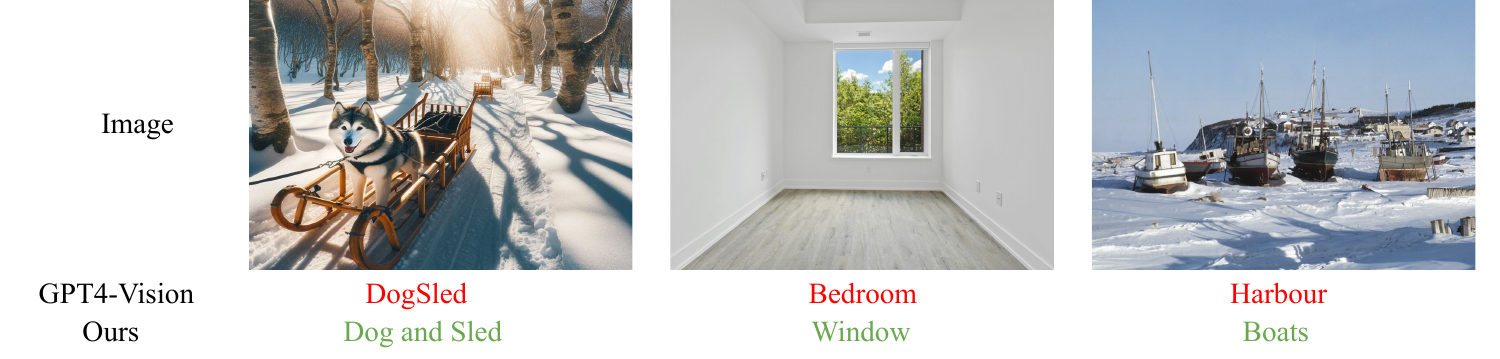}
\caption{Qualitative examples } \label{qualitative}
\vspace{-15pt}
\end{figure}

Our method without image conditioning in Tab. \ref{tab:perfromance_object} surpasses all baselines, including GPT and KG baselines. On other hand, image-conditioned method in Tab. \ref{tab:performance_image} exhibits slightly lower performance compared to GPT on the smaller test set. However, we achieve higher accuracy on the complete set and approach human-level performance. The results highlight the effectiveness of our approach in tackling compound concept classification tasks. Additionally,
% we showcase qualitative results, particularly focusing on cases where ChatGPT 4 fails while our method achieves nearly 100\% accuracy. These results further underscore the capabilities of our approach.
we showcase scenarios where GPT4-Vision fails in Fig. \ref{qualitative}, contrasting with our near-perfect accuracy. Our interpretable approach excels in complex scenarios, leveraging scene and knowledge graphs to outperform GPT4-Vision in challenging tasks.
\vspace{-12pt}
\section{Conclusion}
\vspace{-10pt}
In summary, this work presents a novel approach for compound concept predictions utilizing scene and knowledge graphs. 
Through our method, we propose an effective approach to merge spatial information with general knowledge inference and demonstrate its efficacy compared to a set of state-of-the-art baselines. 
In future work, we intend to expand our approach to video understanding, thus, incorporating spatio-temporal reasoning with knowledge graphs.

\vspace{-10pt}
\bibliographystyle{splncs04}
\bibliography{refs}
% \end{thebibliography}
\end{document}